\documentclass[letterpaper, 10 pt, conference]{ieeeconf}  % Comment this line out if you need a4paper

\IEEEoverridecommandlockouts                              % This command is only needed if 
\usepackage{graphics} % for pdf, bitmapped graphics files
\usepackage{epsfig} % for postscript graphics files
\usepackage{amsmath} % assumes amsmath package installed
\usepackage{amssymb}  % assumes amsmath package installed
\usepackage{url}
\usepackage{flushend}

\usepackage{xcolor}
\usepackage{subcaption}
\usepackage{mwe}

\title{\LARGE \bf
Robust Contact State Estimation in Humanoid Walking Gaits}

\author{Stylianos Piperakis$^{1}$, Michael Maravgakis$^{1}$,  Dimitrios Kanoulas$^{2}$, and Panos Trahanias$^{1}$% <-this % stops a space
\thanks{*This work has been partially supported by the following funded projects: Greek National grand MIS 5030440 vipGPU, EU H2020 TERRINet, UKRI Future Leaders Fellowship [MR/V025333/1] (RoboHike)}% <-this % stops a space
\thanks{$^{1}$Stylianos Piperakis, Michael Maravgakis, and Panos Trahanias are with the Institute of Computer Science, Foundation
for Research and Technology - Hellas (FORTH), Heraklion, Greece. {\tt\small \{spiperakis, maravgakis,trahania\}@ics.forth.gr}}%
  \thanks{$^{2}$Dimitrios Kanoulas is with the Computer Science Department, University College London, UK. {\tt \small d.kanoulas@ucl.ac.uk}}}%
\begin{document}

\maketitle
\thispagestyle{empty}
\pagestyle{empty}

%%%%%%%%%%%%%%%%%%%%%%%%%%%%%%%%%%%%%%%%%%%%%%%%%%%%%%%%%%%%%%%%%%%%%%%%%%%%%%%%
\begin{abstract}
In this article, we propose a deep learning framework that provides a unified approach to the problem of leg contact detection in humanoid robot walking gaits.  Our formulation accomplishes to accurately and robustly estimate the contact state probability for each leg (i.e., stable or slip/no contact). The proposed framework employs solely  proprioceptive sensing and although it relies on simulated ground-truth contact data for the classification process,  we demonstrate that it generalizes across varying friction surfaces and different legged robotic platforms and, at the same time, is readily transferred from simulation to practice.  The framework is  quantitatively and qualitatively assessed in simulation via the use of ground-truth contact data and is contrasted against state-of-the-art methods with an ATLAS, a NAO, and a TALOS humanoid robot. Furthermore, its efficacy is demonstrated in base estimation with a real TALOS humanoid. To reinforce further research endeavors, our implementation is offered as an open-source ROS/Python package, coined Legged Contact Detection (LCD).
\end{abstract}

%%%%%%%%%%%%%%%%%%%%%%%%%%%%%%%%%%%%%%%%%%%%%%%%%%%%%%%%%%%%%%%%%%%%%%%%%%%%%%%%
\section{INTRODUCTION}
Humanoid robot locomotion can be regarded as a sequence of foot contacts that the humanoid experiences with the environment. To this end, accurate and robust foot contact detection entails a vital role in locomotion control~\cite{IHMC, Herzog2016, NeunertNMPC}, gait planning~\cite{Aceituno2018, Winkler2018, Hereid2018}, base state estimation~\cite{Bloesch, Rotella, Raghavan2018, PiperakisIROS19} and Center of Mass (CoM) estimation~\cite{RotellaMomentum, PiperakisEst, PiperakisRAL}.  Therefore, to achieve truly agile and dexterous locomotion, the leg contact status must be accurately estimated.  Nevertheless, this topic remains largely unexplored in humanoid robotics research with some notable exceptions.

Contemporary contact detection approaches can be broadly categorized into two groups: a) approaches that directly employ the measured ground reaction wrenches,  and b)  approaches that incorporate kinematics and dynamics to estimate the Ground Reaction Forces (GRFs) in order to infer the contact status.

Fallon et al.~\cite{Fallon} utilized a Schmitt-Trigger method to classify the measured vertical GRFs from Force/Torque (F/T) sensors in the feet of an Atlas humanoid robot to multiple contact states and determine which leg should be used for state estimation~\cite{Kuindersma}.  This method was also adopted in~\cite{PiperakisRAL} with the pressure sensors of a NAO humanoid.  Bloesch et al.~\cite{BloeschSLIP} employed the binary contact sensors in the feet of the StarlETH quadruped to detect contact and took advantage of the support leg kinematics constraints for updating an Unscented Kalman filter.  Rotella et al.~\cite{RotellaContact} employed F/T and IMU measurements from the feet of a humanoid robot in clustering with a fuzzy c-means algorithm to independently estimate  the contact probabilities for each one of the six leg Degrees of Freedom (DoFs). Moreover, the authors used the obtained contact probabilities in base estimation by adapting accordingly the kinematic measurement uncertainty. 

On the contrary, Ortenzi et al.~\cite{Ortenzi} proposed an approach to estimate the contact constraints the robot experiences with the environment based only on joint position measurements.  Hwangbo et al.~\cite{Hwangbo} introduced a one-dimensional probabilistic framework with a Hidden Markov Model that takes advantage of kinematics, differential kinematics, and dynamics to estimate the contact state. This approach does not rely on F/T sensors, but effectively exploits joint position, velocity, and torque measurements to estimate the GRF.  In~\cite{Neunert}, the contact status of a quadruped robot is inferred from the GRFs, by thresholding the robot dynamics.  Recently, Camurri et al.~\cite{Camurri} demonstrated a supervised learning framework that employs logistic regression to estimate the contact probabilities for quadruped robots.  This one-dimensional classifier utilized the estimated GRF from dynamics, joint position, and torque measurements to encode different GRF thresholds for different type of gaits.  Nevertheless, to perform the training, the ground-truth base velocity is needed. Similarly, Lim et al. ~\cite{lin2021legged} developed a deep learning-based contact estimator that uses proprioceptive sensory data as input and classifies the individual contacts as a binary state. Despite the fact that the results indicate high classification accuracy, the framework is coupled with a specific robot and controller and it is unable to generalize to different platforms without new ground-truth labeled data.

Most of the aforementioned approaches determine whether a specific leg experiences contact with the environment or not. Recent works try to estimate directly the gait-phase during locomotion. Towards this direction, in~\cite{Bledt2018}, a linear Kalman Filter is utilized to estimate each leg state (swing or contact) for quadruped robots. The latter employs Gaussian probabilistic models for the contact forces and the terrain's ground height to infer the gait-phase. Although a very high estimation accuracy is recorded, the scheme relies on prior knowledge from pre-planned contacts and gait-phases and thus directly couples the control and estimation processes.  Recently, we proposed 
an unsupervised learning framework, coined Gait-phase Estimation Module (GEM)~\cite{PiperakisGEM}, that takes advantage of linear dimensionality reduction with PCA on proprioceptive sensing and clustering in the latent space, to infer the gait-phase probabilities. A high accuracy for all three gait-phases was demonstrated with a simulated Valkyrie robot, but only statically stable walking was examined.

\subsection{Contribution}
In this article, we propose a deep learning framework based on proprioception, specifically an F/T and an IMU sensor in each leg, to determine the contact state probabilities, namely stable or slip/no contact probabilities for dynamic walking gaits over variable friction surfaces that can further benefit the legged locomotion problem.  Our contribution to the state-of-the-art regards:
\begin{itemize}
    \item A unified approach for contact detection.  We demonstrate that a model trained with walking gaits over a specific friction coefficient, generalizes to a very large range of frictions. Additionally, the model also generalizes to different robotic platforms.  Although our model is trained with the ATLAS robot, the same model  provides  highly accurate contact estimation in NAO and TALOS walking gaits.
    \item A framework that relies solely on proprioceptive sensing that is readily available in contemporary humanoids.
    \item A demonstration that, although the model is trained in simulation with the ground-truth contact states as labels, it can be readily employed to infer the contact state with a real TALOS humanoid.
    \item A framework that has been extensively evaluated against state-of-the-art approaches in contact estimation and it's efficiency is demonstrated both in simulation and real robot experiments.
    \item The release of an open-source module implementation in ROS/Python, named Legged Contact Detection (LCD) module~\cite{LCD}.
\end{itemize}

Notice that in~\cite{RotellaContact}, IMUs in the legs are also considered. In this case a shortcoming is that $12$ fuzzy c-means clustering models must be individually trained, one for each of the three translational and rotational DoFs of both legs, to collectively estimate the leg contact states. Accordingly, in~\cite{Camurri}, two supervised classifiers are trained, one for each leg with the ground-truth base velocities recorded by a motion capture system as labels.  In the gait-phase estimation framework of~\cite{PiperakisGEM}, the latent dimension must be pre-specified to perform  dimensionality reduction either with PCA or autoencoders. The latter highly depends on the pace and speed of the gait.

% In this work we follow a radically different approach.  Firstly, we consider data to be received by F/T and IMU sensors in the robot's feet.  Secondly, we explore contact model training using a single robot (e.g., ATLAS) in simulation for just a single surface friction value.  Our data collection and training method: 1) achieves highly accurate and robust surface contact detection, 2) generalizes the contact estimation to surfaces of friction that were not in the training dataset, 3) generalizes well in different simulated robotic platforms (from light to heavy weighted), and 4) can be applied directly to real-robots.  We, thus, propose a novel friction- and robot- invariant method for accurate and robust contact estimation, based only on GRF and IMU sensory data.  This allows of course further direct application to other types of feet, such as point-contact ones.

The current article is organized as follows: Section~\ref{sec:TDA} presents a physical interpretation to the particular choice of training data. In Section~\ref{sec:LCD} the deep learning contact detection  framework is presented. Subsequently, the proposed framework is quantitatively and qualitatively  assessed both in simulation and real-world experiments in Section~\ref{sec:experiments}.  Finally, Section~\ref{sec:conclusions} concludes the article and outlines potential future work.
%%%%%%%%%%%%%%%%%%%%%%%%%%%%%%%%%%%%%%%%%%%%%%%%%%%%%%%%%%%%%%%%%%%%%%%%%%%%%%%%

\section{Training Data Acquisition}
\label{sec:TDA}
Training data is an important aspect of machine learning. Instead of blindly employing all available sensory data in a training session, we provide a physical interpretation to a particular choice of features that are directly correlated to the contact state. In the following, we assume that the robot is equipped with proprioceptive sensing that is commonly available in humanoids nowadays, namely, F/T and IMUs sensors in the legs. The datasets used for LCD training and testing are released in~\cite{LCD}.

\subsection{Contact State in the Centroidal Dynamics}
The centroidal dynamics of a humanoid during locomotion can be described by the Newton-Euler equations:
\begin{align}
m(\ddot{\boldsymbol{c}} + \boldsymbol{g}) &=  \sum_i \boldsymbol{f}_i \label{eq:Forces}\\ 
m\boldsymbol{c} \times (\ddot{\boldsymbol{c}} + \boldsymbol{g}) + \dot{\boldsymbol{L}} &	=  \sum_i \boldsymbol{s}_i \times\boldsymbol{f}_i + \boldsymbol{\tau}_i \label{eq:Moments}
\end{align}
where $\boldsymbol{c}$ and  $\ddot{\boldsymbol{c}}$ are the CoM position and acceleration,  $\dot{\boldsymbol{L}}$ is angular momentum rate around the CoM, $\boldsymbol{f}_i$ and $\boldsymbol{\tau}_i$ are  the Ground Reaction Forces (GRFs) and Torques (GRTs), $\boldsymbol{s}_i$ are the contact points, $\boldsymbol{g}$ is the gravity vector, and $m$ is the robot's mass.

Subsequently, in order for a leg to maintain contact and neither slip nor rotate, the friction constraints must apply:
\begin{align}
\sqrt{ (f^{x}_i)^2 +  (f^{y}_i)^2} &\leq \mu^{x,y} f^{z}_i \label{eq:frictionCon}\\
-\tau^y_i/f_i^z &\leq p^x \\
\tau^x_i/f^z_i &\leq p^y \\
|\tau^z_i| &\leq \mu^z f_i^z \label{eq:tauCon}
\end{align}
where  $\boldsymbol{p}$  is the center of pressure and $\mu^{x,y},\mu^{z}$ are the planar and rotational contact friction coefficients, respectively.

\begin{figure*}[ht]
    \centering
    \includegraphics[width =1.15\columnwidth]{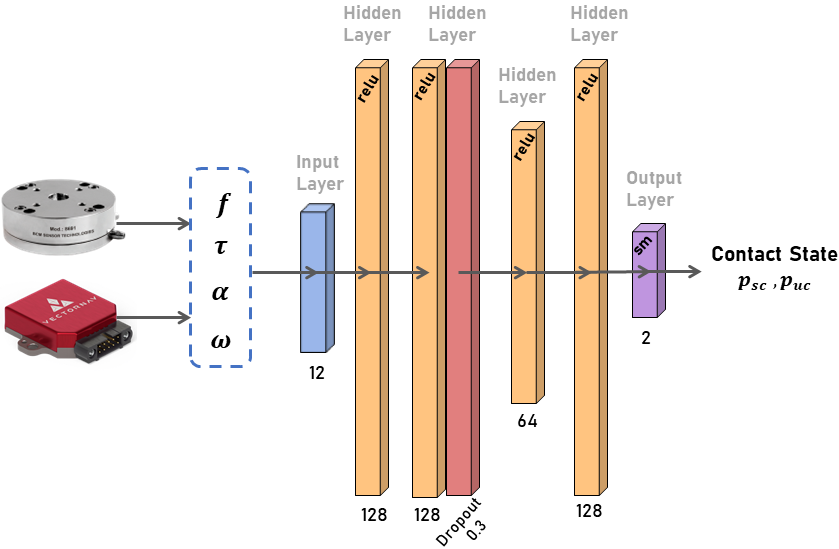}
    \caption{LCD Deep Learning Architecture.}
    \label{fig:LCD}
\end{figure*}

% \begin{figure*}[t!]
%     \centering
%     \includegraphics[width=\textwidth]{robots.png}
%     \caption{Experiments ( Not final image ) }
%     \label{fig:Force}
% \end{figure*}
    % \begin{figure*}
    %     \centering
    %     \begin{subfigure}[b]{0.475\textwidth}
    %         \centering
    %         \includegraphics[width=\textwidth, height = 4.5cm]{atlas.png}
    %         \caption[Network2]%
    %         {{\small Simulated ATLAS experimental setup}}    
    %         \label{fig:mean and std of net14}
    %     \end{subfigure}
    %     \hfill
    %     \begin{subfigure}[b]{0.475\textwidth}  
    %         \centering 
    %         \includegraphics[width=\textwidth, height = 4.5cm]{nao.png}
    %         \caption[]%
    %         {Simulated NAO experimental setup}    
    %         \label{fig:mean and std of net24}
    %     \end{subfigure}
    %     \vskip\baselineskip
    %     \begin{subfigure}[b]{0.475\textwidth}   
    %         \centering 
    %         \includegraphics[width=\textwidth,height = 4.5cm]{talos.png}
    %         \caption[]%
    %         {Simulated TALOS experimental setup}    
    %         \label{fig:mean and std of net34}
    %     \end{subfigure}
    %     \hfill
    %     \begin{subfigure}[b]{0.475\textwidth}   
    %         \centering 
    %         \includegraphics[width=\textwidth,height = 4.5cm]{TALOS.png}
    %         \caption[]%
    %         {Real TALOS robot}    
    %         \label{fig:mean and std of net44}
    %     \end{subfigure}
    %     \caption[]
    %     {Experimental setup for variable friction surfaces. {\color{red} Figure cited in text AFTER fig 2.}} 
    %     \label{fig:experiments}
    % \end{figure*}

As evident there is a direct correlation of the contact points and, thus, the contact state with the ground reaction wrenches and the centroidal dynamics.
Although we can measure the left and right leg contact wrenches ${}^l\boldsymbol{f}_l,{}^l\boldsymbol{\tau}_l$ and ${}^r\boldsymbol{f}_r,{}^r\boldsymbol{\tau}_r$ with F/T sensors in the local leg frames and also compute the CoM velocity ${}^b\boldsymbol{\dot{c}}$ and angular momentum rate ${}^b\boldsymbol{\dot{L}}$ in the base frame with kinematics, friction depends on the environment and prohibits the analytical derivation of the contact state.

\subsection{Contact State in the  Leg Kinematics}
The contact state is also directly linked to the leg kinematics namely, the left and right leg spatial linear and angular velocities ${}^{l} \boldsymbol{v}_l, {}^{l} \boldsymbol{\omega}_l$ and ${}^{r} \boldsymbol{v}_r, {}^{r} \boldsymbol{\omega}_r$. More specifically, for the left  leg to experience a stable contact with the environment and not slip in the tangential directions, the following conditions must apply:
\begin{align}
    {}^l f_l^z &> 0 \\
    {}^l v_l^x &= 0 \label{eq:velCon}\\
    {}^l v_l^y &= 0 \\
    {}^l\omega_l^z &=0
\end{align}

Furthermore, when the leg is stationary on the ground and is not breaking the contact by lifting nor rotating, then:
\begin{align}
    {}^l v_l^z &= 0 \\
    {}^l \omega_l^x &= 0  \\
    {}^l \omega_l^y &= 0 \label{eq:omegaCon}
\end{align}
Accordingly, the same conditions apply to the right leg.

In the above, the spatial rotational velocities ${}^{l,r}\boldsymbol{\omega}_{l,r}$ can be directly measured with an IMU attached to the foot links. On the contrary, the spatial linear velocities ${}^{l,r}\boldsymbol{v}_{l,r}$ cannot be measured and must be estimated. To avoid introducing correlations between the base and the contact state estimation, we employ the leg spatial linear accelerations ${}^{l,r} \alpha_{l,r}$, which can also be measured by the leg IMUs.

Although the spatial linear accelerations carry similar dynamic information about the leg contact state they fail to capture the case where the legs are slipping with constant spatial linear velocity, e.g., when stepping on ice. Nevertheless, the latter is not yet a realistic case for modern legged robots and will not be considered in this study.

\section{Robust Contact Estimation with Deep Learning}
\label{sec:LCD}

To accurately infer the leg's contact state we devised a supervised learning framework, termed Legged Contact Detection (LCD), depicted in Figure \ref{fig:LCD}. The data employed for the training procedure were the leg F/T measurements, namely ${}^{l}\boldsymbol{f}_{l}, {}^{l}\boldsymbol{\tau}_{l}$ and the leg IMU data, namely ${}^{l}\boldsymbol{\omega}_{l}, {}^{l}\boldsymbol{\alpha}_{l}$, as measured in the local leg frame. A single model is trained with the left and right leg F/T and IMU data and is used to infer the contact states for both legs.

% \begin{figure*}[ht]
%     \centering
%     \includegraphics[width =0.95\columnwidth]{LCD_arch.png}
%     \caption{LCD Deep Learning Architecture. {\color{red} Make this larger}}
%     \label{fig:LCD}
% \end{figure*}
  \begin{figure*}
        \centering
        \begin{subfigure}[b]{0.475\textwidth}
            \centering
            \includegraphics[width=\textwidth, height = 4.5cm]{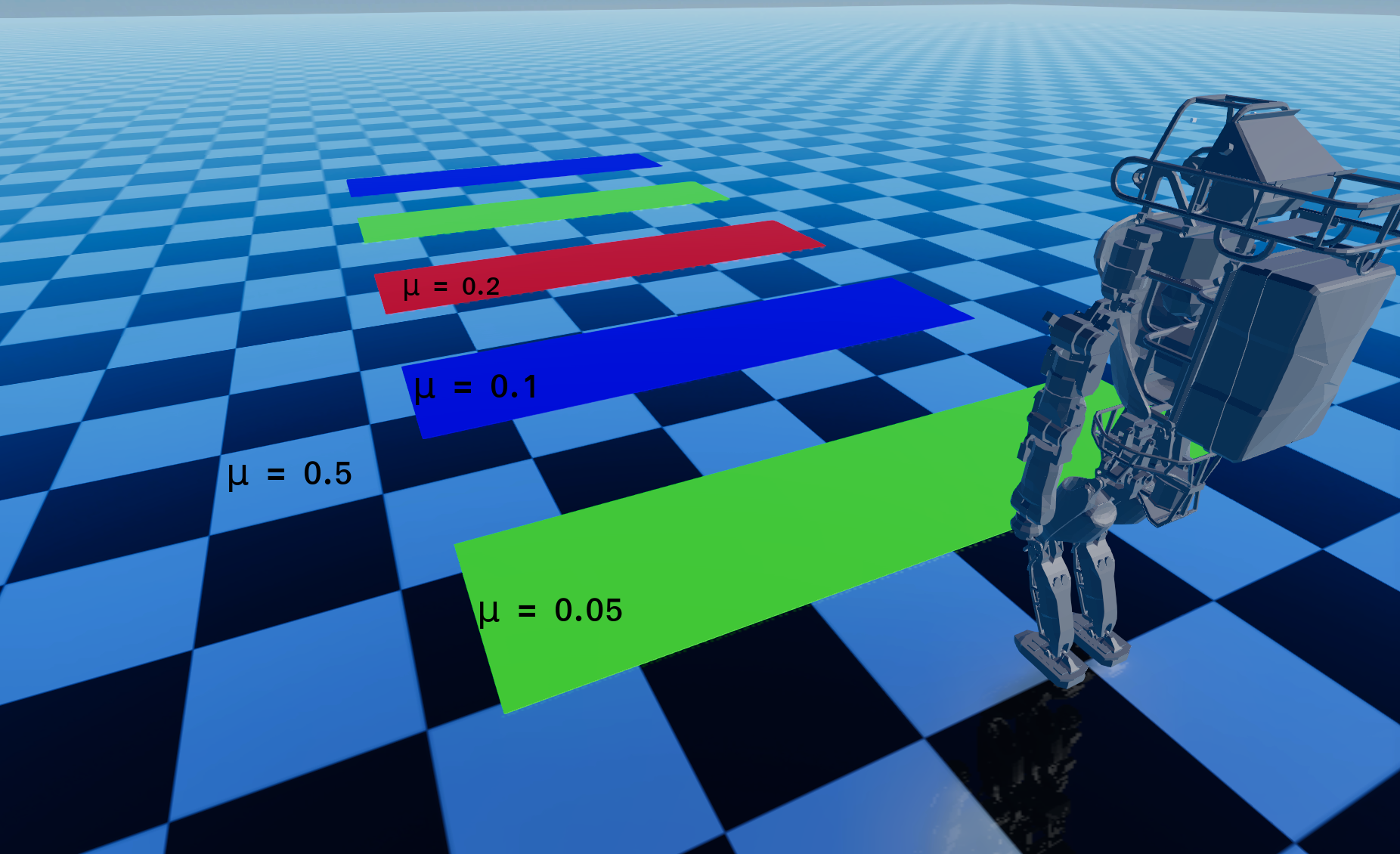}
            \caption[Network2]%
            {{\small Simulated ATLAS experimental setup}}    
            \label{fig:mean and std of net14}
        \end{subfigure}
        \hfill
        \begin{subfigure}[b]{0.475\textwidth}  
            \centering 
            \includegraphics[width=\textwidth, height = 4.5cm]{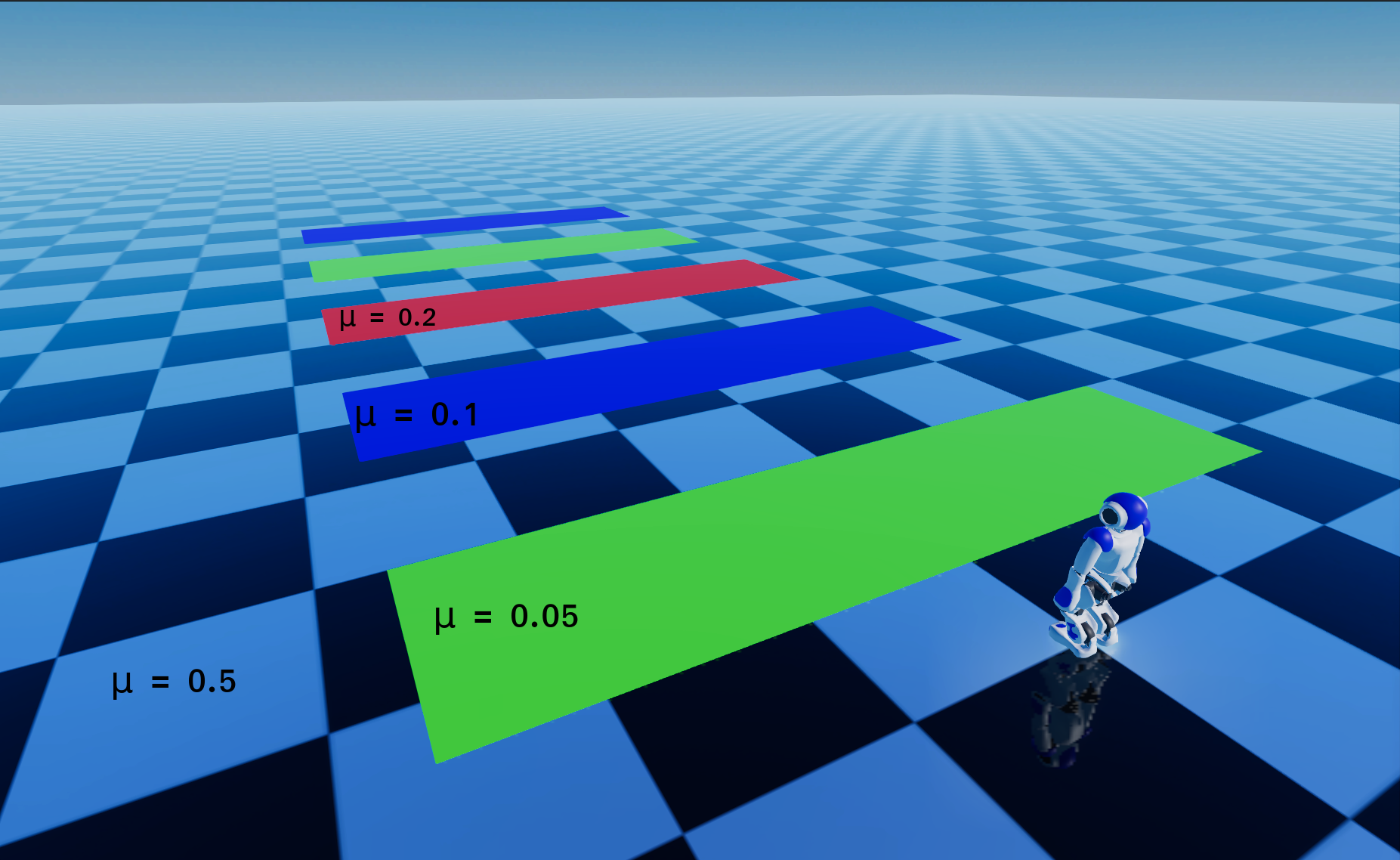}
            \caption[]%
            {Simulated NAO experimental setup}    
            \label{fig:mean and std of net24}
        \end{subfigure}
        \vskip\baselineskip
        \begin{subfigure}[b]{0.475\textwidth}   
            \centering 
            \includegraphics[width=\textwidth,height = 4.5cm]{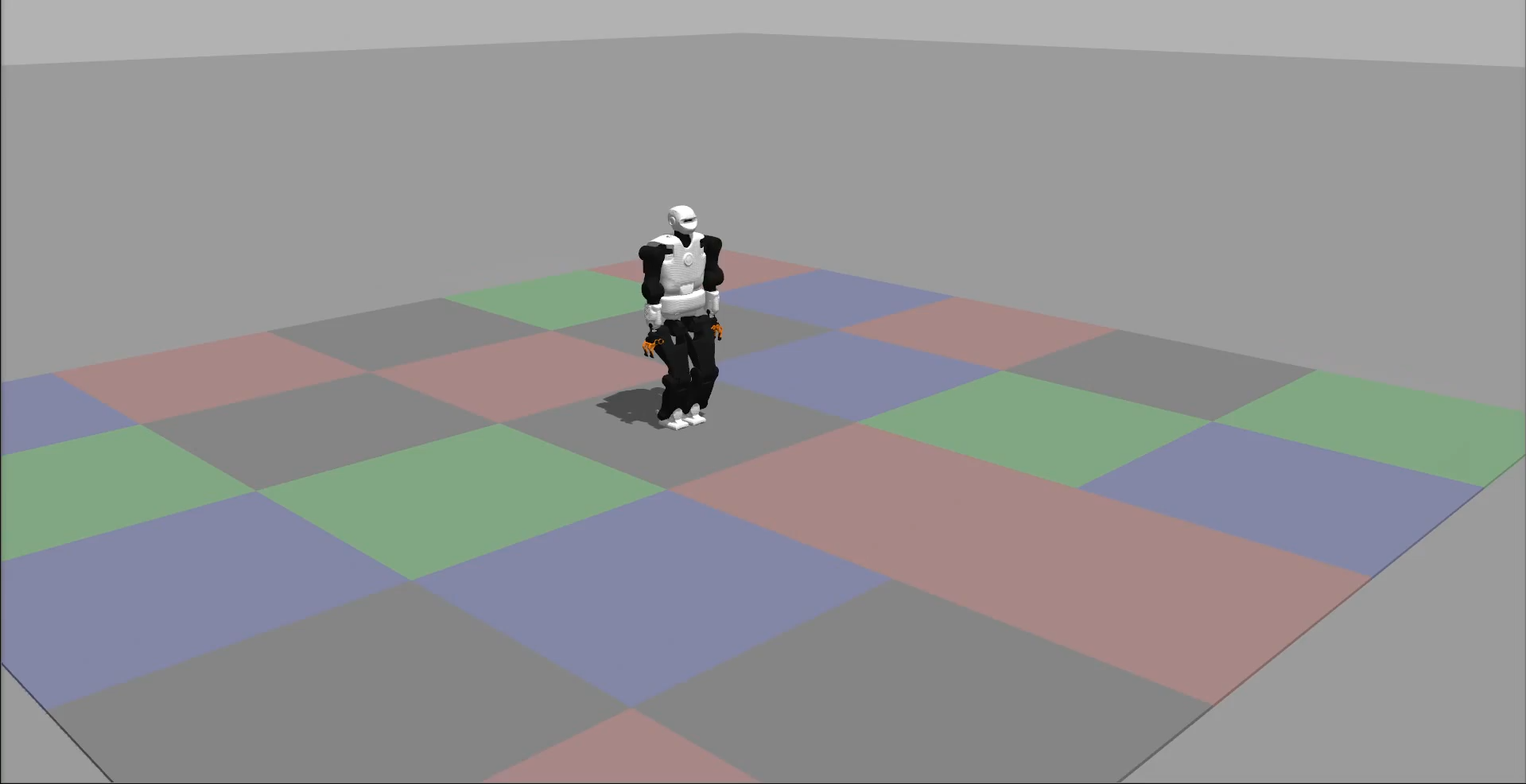}
            \caption[]%
            {Simulated TALOS experimental setup}    
            \label{fig:mean and std of net34}
        \end{subfigure}
        \hfill
        \begin{subfigure}[b]{0.475\textwidth}   
            \centering 
            \includegraphics[width=\textwidth,height = 4.5cm]{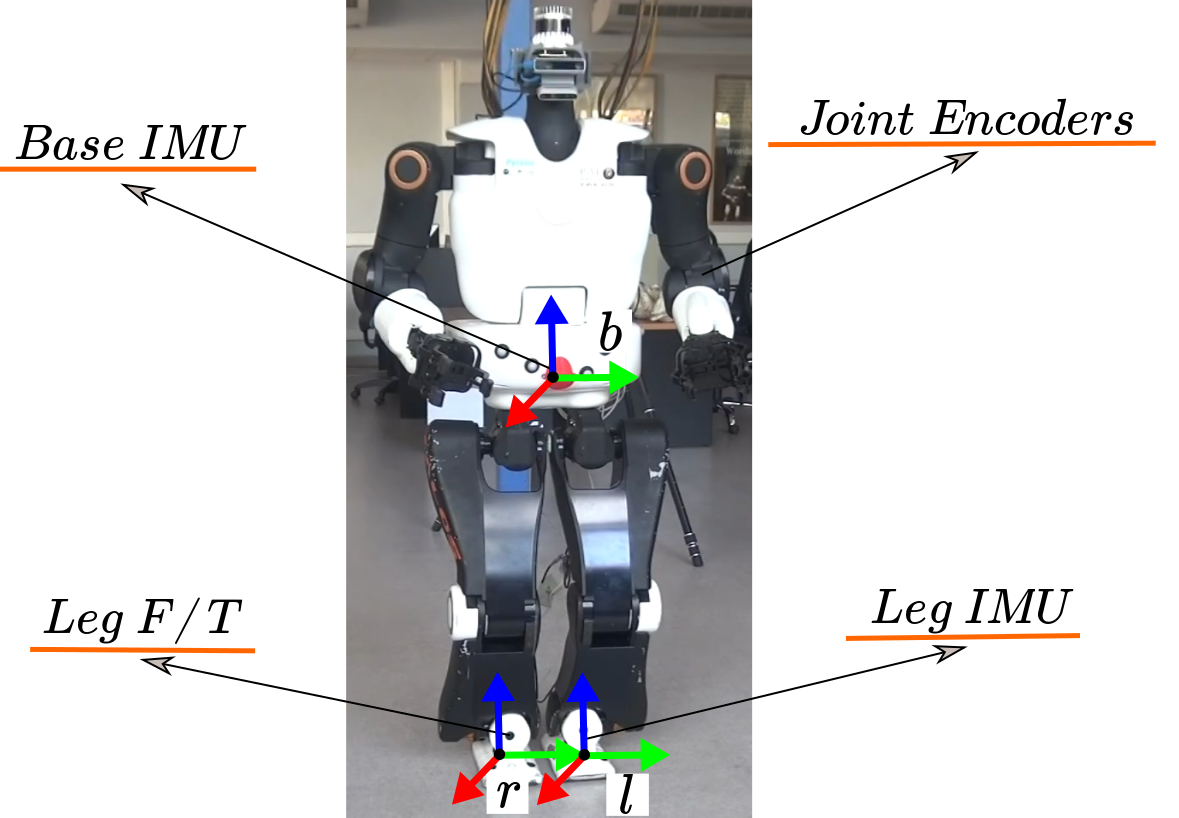}
            \caption[]%
            {Real TALOS robot}    
            \label{fig:mean and std of net44}
        \end{subfigure}
        \caption[]
        {Experimental setup for variable friction surfaces.} 
        \label{fig:experiments}
    \end{figure*}

\subsection{Preprocessing}

For the F/T measurements the following model was considered:
\begin{align}
    {}^l \boldsymbol{f}_l &= {}^l   \boldsymbol{\bar{f}}_l + \boldsymbol{b}_f +\boldsymbol{w}_f\\
    {}^l \boldsymbol{\tau}_l &=  {}^l \boldsymbol{\bar{\tau}}_l + \boldsymbol{b}_\tau + \boldsymbol{w}_\tau
\end{align}
where the ${}^l\boldsymbol{\bar{f}}_l$, ${}^l\boldsymbol{\bar{\tau}}_l$ are the true GRFs and GRTs, $\boldsymbol{b}_f$, $\boldsymbol{b}_\tau$ and $\boldsymbol{w}_f$, $\boldsymbol{w}_\tau$ are the F/T measurement biases and zero-mean Gaussian  noises, respectively. 

Similarly, for the IMU measurements the following model was employed:
\begin{align}
    {}^l \boldsymbol{\alpha}_l &= {}^l   \boldsymbol{\bar{\alpha}}_l + {}^l \boldsymbol{R}_w \boldsymbol{g}  + \boldsymbol{b}_\alpha + \boldsymbol{w}_\alpha\\
    {}^l \boldsymbol{\omega}_l &=  {}^l \boldsymbol{\bar{\omega}}_l + \boldsymbol{b}_\omega + \boldsymbol{w}_\omega
\end{align}
where the ${}^l \boldsymbol{\bar{\alpha}}_l$, ${}^l \boldsymbol{\bar{\omega}}_l$ are the true linear acceleration and angular velocity, ${}^l \boldsymbol{R}_w$ is the rotation from the world to the left leg frame, $\boldsymbol{g}$ is the gravity vector, $\boldsymbol{b}_\alpha$, $\boldsymbol{b}_\omega$ and $\boldsymbol{w}_\alpha$, $\boldsymbol{w}_\omega$ are the IMU measurement biases and zero-mean Gaussian noises. Evidently, the
same models apply for the right leg F/T and IMU measurements.

Initially, the biases for all measurements have been removed, including the gravity constant in the linear acceleration measurements. Next, measurements exceeding $3\sigma$ were identified as outliers and eliminated from the dataset. 

All data have been normalized in each dimension with their maximum value to avoid large scale measurements such as the vertical GRF  ${}^{l}f_{l}^z$  dominating the learning procedure. Subsequently, the absolute value was taken since slip is bidirectional and does not depend on measurement signs.
Moreover, all data have been synchronized and downsampled to $100Hz$ since the contact state commonly changes when the robot takes a step which contemporary humanoids accomplish with a slower rate, e.g., $1-2Hz$.

\subsection{Architecture}
The LCD network, illustrated in Figure~\ref{fig:LCD},consists of $2$ hidden layers with $128$ neurons followed by a $30\%$ dropout layer to prevent overfitting. Subsequently, two more hidden layers were added with $64$ and $128$ neurons, respectively, to feed an output layer of $2$ units, one for each contact probability, namely stable contact or unstable/no contact. For all hidden layers the ReLU activation was used, while for the output layer the sigmoid was employed to guarantee that the output is a valid probability. The overall architecture was determined experimentally while aiming to maximize the accuracy of the classifier on data acquired from other robotic platforms than the one employed for training. Hyperparameter grid search was performed to optimize the efficiency of the network. Overall, LCD was trained for 30 epochs, with a batch size of 16 and the \textit{adam} optimizer.
%\textcolor{red}{followed with a Max Pooling layer of stride $2$ to downsample the data and a Flatten layer are employed}. This part was designed and optimized by hyperparameter grid searching to effectively extract deep contact features from the input data. 
% A final hidden layer of $128$ neurons is used to 

Accordingly, we formulate a supervised classification problem by minimizing the binary cross-entropy loss:
\begin{equation}
L = - \left(y_{sc} \log(p_{sc}) + (1-y_{sc}) \log(1-p_{sc})\right)
\end{equation}
where $p_{sc}$ is the stable contact probability, $p_{uc} = 1-p_{sc}$ is the unstable/no contact probability and $y_{sc}$ is the ground truth stable contact label obtained by evaluating  Eq.~(\ref{eq:frictionCon})~-~(\ref{eq:tauCon}) as well as Eq.~(\ref{eq:velCon})~-~(\ref{eq:omegaCon}) in simulation, as also outlined in the next section.

\section{RESULTS}\label{sec:experiments}
In the current section, we present  quantitative and qualitative results that demonstrate the accuracy and efficacy of the proposed framework both in simulation and real world experiments. LCD was implemented in ROS/Python and is publicly available at~\cite{LCD}. A snapshot of the experimental setup is illustrated in Figure \ref{fig:experiments}. In addition all of our experiments are presented in high resolution at \url{https://youtu.be/csUIadkT7OM}.
% All our experiments are presented in high resolution in the supplementary video material \textcolor{red}{and a snapshot of the setup is illustrated in Figure \ref{fig:experiments}.}

\subsection{Simulation Results}
To conduct a quantitative and qualitative assessment, we employed an ATLAS and a NAO humanoid robot in RaiSim~\cite{raisim} --a high-accuracy multi-contact simulator for articulated robots-- and the TALOS humanoid in Gazebo~\cite{Gazebo}.  Accordingly, to generate walking patterns, we've implemented a robot generic omnidirectional walking motion planning~\cite{LIPMMotion} and a real-time gait stabilization module~\cite{LIPMControl}, both based on the Linear Inverted Pendulum (LIPM) dynamics~\cite{Piperakis,ScaronControl}. Subsequently, to realize the desired trajectories in each humanoid, we've also developed a real-time whole body control module~\cite{wholebodyIK} based on stack of tasks at the velocity level~\cite{OpenSot}. In our formulation, higher priority was given to the desired leg trajectories, then to the desired CoM position and torso orientation, and finally to a standing posture joint configuration task to maintain postural balance. Regarding the feedback of the motion planning, real-time stabilization and whole-body inverse kinematics, the ground-truth values were employed. The latter modules are also released as open-source ROS/C++ packages to reinforce further research endeavors.

\begin{figure*}[!t]
    \centering
    \includegraphics[width=\textwidth]{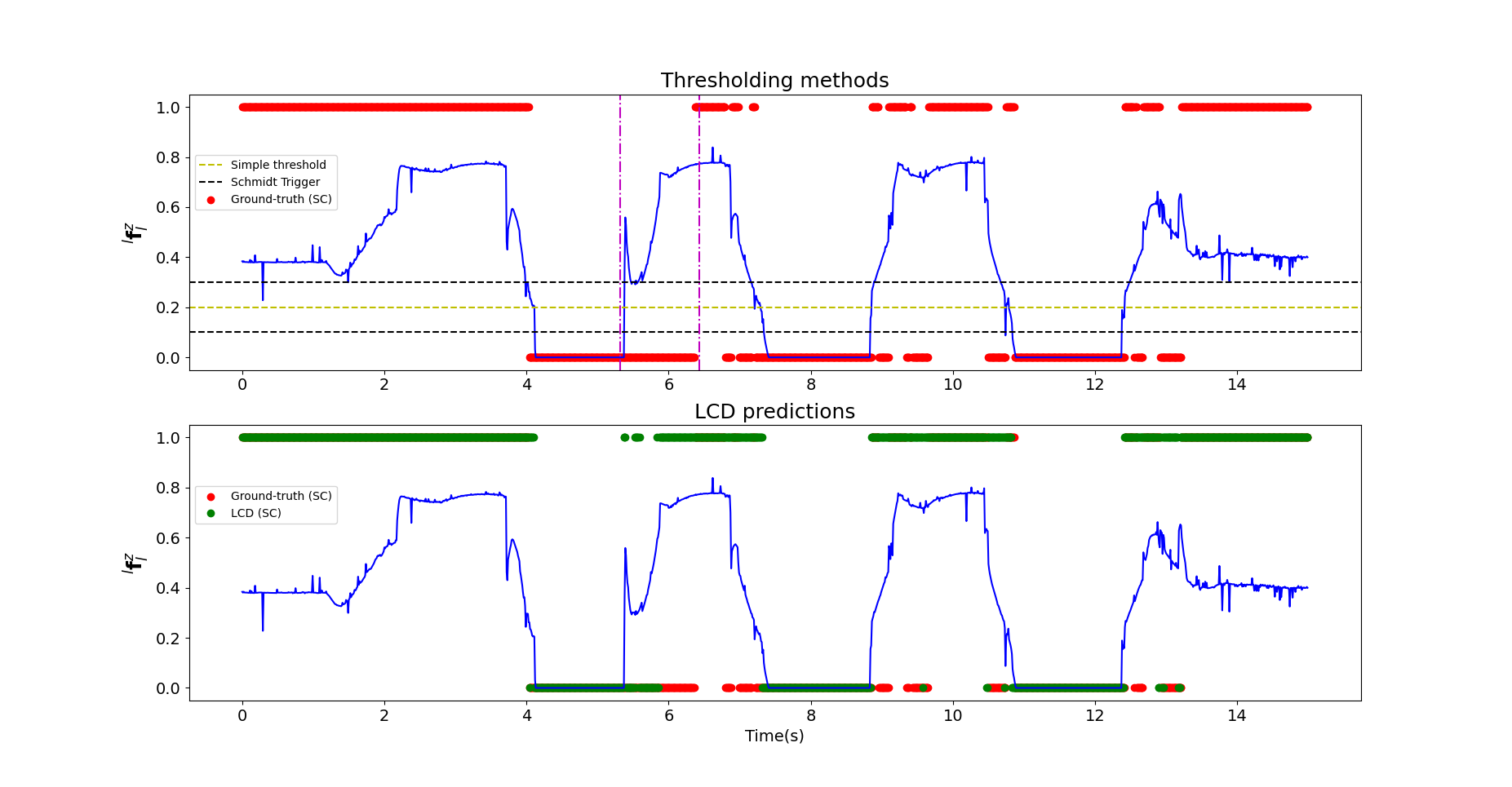}
    \caption{Thresholding and LCD predictions for SC on ATLAS walking gaits with varying friction coefficient surfaces}
    \label{fig:Force}
\end{figure*}

\begin{table}[!b]
\begin{center}
\caption{Simulation noise standard deviations.}
\label{table:NoiseTable}
\begin{tabular}{| c | c | c |}
\hline
  & Continuous & Discrete ($100Hz$) \\ 
 \hline
 $\sigma_{\boldsymbol{\alpha}}$ & $0.0008 m/s^2/\sqrt{Hz}$  & $0.008 m/s^2$ \\  
 $\sigma_{\boldsymbol{\omega}}$ & $0.0005 rad/s/\sqrt{Hz}$ & $0.005 rad/s$ \\
 $\sigma_{\boldsymbol{b_\alpha}}$ & $0.0001m/s^3/\sqrt{Hz}$ & $0.001m/s^3$ \\
 $\sigma_{\boldsymbol{b_\omega}}$ & $0.0006 rad/s^2/\sqrt{Hz}$ &  $0.006 rad/s^2$ \\
 $\sigma_{\boldsymbol{f}}$ & $0.07N/\sqrt{Hz}$ & $0.7N$ \\  
 $\sigma_{\boldsymbol{\tau}}$ & $0.003Nm/\sqrt{Hz}$ & $0.03Nm$ \\
 $\sigma_{\boldsymbol{b_f}}$ & $0.0001N/s/\sqrt{Hz}$ & $0.001N/s$ \\
 $\sigma_{\boldsymbol{b_\tau}}$ & $0.0001Nm/s\sqrt{Hz}$ & $0.001Nm/s$ \\
 \hline
\end{tabular}
\end{center}
\end{table}

Next, we've commanded each robot to continuously walk over multiple surfaces with varying friction coefficients from $0.05$ to $1.2$, for approximately $10$ minutes, to record the needed dataset. Overall, the above sessions resulted in an  average distribution of the labels as follows: $60\%$ for Stable Contact (SC) and $40\%$ for Unstable Contact (UC) ($30\%$  for no contact and $10\%$ for slip).
The legs' IMU and F/T measurements were available at $500Hz$ for ATLAS and TALOS and at $100Hz$ for NAO. In all measurements, i.i.d Gaussian noise was added to provide realistic noise levels according to Table~\ref{table:NoiseTable}.

Subsequently, the LCD model is trained with a $10$ minute omni-directional walking gait via the ATLAS robot and over a $0.2$ friction coefficient surface.  This model is then used to infer the contact state for every walking gait performed with the ATLAS, NAO, and TALOS robot over variable friction surfaces. To compute the necessary training labels, we evaluate Eqs.~(\ref{eq:frictionCon})--(\ref{eq:tauCon}) and Eqs.~(\ref{eq:velCon})--(\ref{eq:omegaCon}) using the ground-truth values at every discrete time instant. If the latter is true, the contact label is characterized as SC, otherwise it is a UC. Note that UC includes the slip and no contact states.

To quantitatively assess the proposed framework in terms of accuracy, we employ several state-of-the-art contact detection methods. More specifically, we have implemented a) the vertical GRF thresholding (T), b) the Schmidt Trigger~\cite{Fallon} (ST), which relies on hysteresis thresholding with a low and a high vertical GRF threshold, and c) the fuzzy c-means (FCM) contact detector~\cite{Rotella}. The first two are binary contact classification methods while the third is a contact probability detector based on leg F/T and IMU data clustering. The thresholds employed for each robot were finely tuned for each dataset to yield the best results, while for FCM the fuzziness parameter was set to $1.2$ and a batch size of $20$ input samples was used for all robots.

\subsection*{Comparison to Thresholding methods}

% \begin{table}[!b]
% \centering
% \caption{Vertical GRF threshold and Schmidt Trigger values.}
% \label{table:Params}
% \begin{tabular}{| c | c | c |}
% \hline
%   & T - Threshold (N) & ST - Low/High Threshold (N) \\ 
%  \hline
%  ATLAS & 512 & 341/853 \\  
%  NAO & 15 & 10/25 \\
%  TALOS & 268 & 179/446\\
%  \hline
% \end{tabular}
% \end{table}

\begin{table*}[ht]
\centering
\caption{LCD evaluation on variable friction datasets}
\label{table:Eval}
\begin{tabular}{|l|l|l|l|l|l|l|l|}
\hline
& \multicolumn{2}{|c|}{LCD} & \multicolumn{2}{|c|}{Simple Threshold} & \multicolumn{2}{|c|}{Schmidt Trigger}\\ 
Dataset  & SC(\%) & UC(\%) & SC(\%)      & UC(\%)      & SC(\%)       & UC(\%)       \\ 
\hline
ATLAS single friction(15k) & 97 & 96 & 97 & 93 & 96 & 92             
\\
ATLAS mixed frictions(50k) & 96 & 84 & 95 & 79 & 94 & 81      
\\
NAO mixed frictions(15k) & 96 & 80 & 94   & 73 & 89 & 75       
\\
TALOS mixed frictions(50k) & 92 & 70 & 98 & 64 & 99 & 65   
\\    
\hline
\end{tabular}
\end{table*}

% Michalis 
The quantitative results (Table~\ref{table:Eval}) from the comparison between LCD, T, and ST indicate that LCD outperforms every thresholding model in identifying the UC state. Although the difference is between 3-7\%, this is rather significant because slip occurs rarely and for a short period of time and, thus, the no contact class dominates in size the UC labels. Figure~\ref{fig:Force} demonstrates how the vertical force of the left foot varies during gait and the ground truth labels for SC (1.0) and UC (0.0). The latter presents two rows whereby the top row refers to the basic thresholding methods, namely T and ST, and the bottom row presents our own results.
The depicted gait pattern is extracted from the ATLAS robot while walking  on surfaces with varying friction coefficients. Each peak represents a step, more specifically the first step is on terrain with $\mu = 0.5$, where $\mu$ stands for the terrain-foot friction coefficient. Similarly,  the second peak regards a case with $\mu=0.05$\footnote{A value of $\mu = 0.05$ refers to walking on almost ice-like surfaces.}, and the third and forth  peaks refer to $\mu = 0.1$ and $\mu =0.5$, respectively. Note that the gait phase in the initial part of the first step (peak) and the final part of the last step is Double Support. It is interesting to observe that during the second step, although the robot has transferred its weight to perform the next step, the foot is slipping and hence T and ST are misclassifying the corresponding data points (purple region) since ${}^lF^z_{l}$ is greater than the threshold. On the contrary, this is not the case with the proposed LCD framework, which identifies the UC state of the gait. Similar observations also hold true for the subsequent steps that are illustrated in the same figure.

% \begin{table*}[ht]
% \centering
% \caption{LCD evaluation on variable friction datasets}
% \label{table:Eval}
% \begin{tabular}{|l|l|l|l|l|l|l|l|}
% \hline
% & \multicolumn{2}{|c|}{LCD} & \multicolumn{2}{|c|}{Simple Threshold} & \multicolumn{2}{|c|}{Schmidt Trigger}\\ 
% Dataset  & SC(\%) & UC(\%) & SC(\%)      & UC(\%)      & SC(\%)       & UC(\%)       \\ 
% \hline
% ATLAS single friction & 97 & 96 & 97 & 93 & 96 & 92             
% \\
% ATLAS mixed frictions & 96 & 84 & 95 & 79 & 94 & 81      
% \\
% NAO mixed frictions & 96 & 80 & 94   & 73 & 89 & 75       
% \\
% TALOS mixed frictions & 92 & 70 & 98 & 64 & 99 & 65   
% \\    
% \hline
% \end{tabular}
% \end{table*}

\subsection*{Comparison to unsupervised learning}
\begin{figure}[ht]
    \centering
    \includegraphics[width=.95\columnwidth]{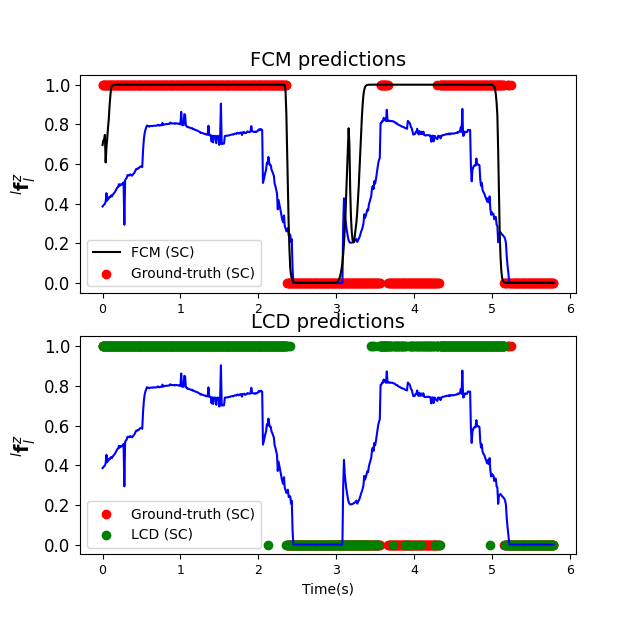}
    \caption{Unsupervised learning and LCD predictions on normal and low friction gaits.}
    \label{fig:FCM_LCD}
\end{figure}

In Figure~\ref{fig:FCM_LCD} we demonstrate a qualitative comparison between unsupervised learning (FCM) and the proposed model (LCD) on the same gaits. The top graph illustrates the probability of stable contact ($p_{sc}$) as computed by the FCM versus the ground truth labels.  FCM accurately predicts the first step and, although it recognises the instability at the beginning of the second step, it quickly converges to the incorrect label. On the other hand, the bottom graph illustrates the predictions of LCD which successfully captures most of the data samples classified as UC (0.0) but also SC (1.0) according to the ground truth labels.

\subsection*{LCD with feature reduction}
In order to test the robustness of LCD and its transferability to point feet robotic platforms (such as quadrupeds), we removed all the F/T measurements from the training dataset except the vertical force $F_z$. Next, we trained the model by using only $F_z$ and IMU measurement. After training, the model was able to make successful predictions on test datasets with different but not extremely low friction coefficients, as shown in Table~\ref{table:leggedresults}.

\begin{table}[!b]
\begin{center}
\caption{LCD performance with reduced features}
\label{table:leggedresults}
\begin{tabular}{| c | c | c |}
\hline
 dataset & SC(\%) & UC(\%) \\ 
 \hline
 ATLAS, $\mu = 0.5$ & 91  & 98 \\  
 ATLAS  $\mu = 0.4$ & 92 &  99 \\
 \hline
\end{tabular}
\end{center}
\end{table}

\subsection{Experimental Results: Application to Base Estimation}
\begin{figure}
    \centering
    \includegraphics[width=\columnwidth]{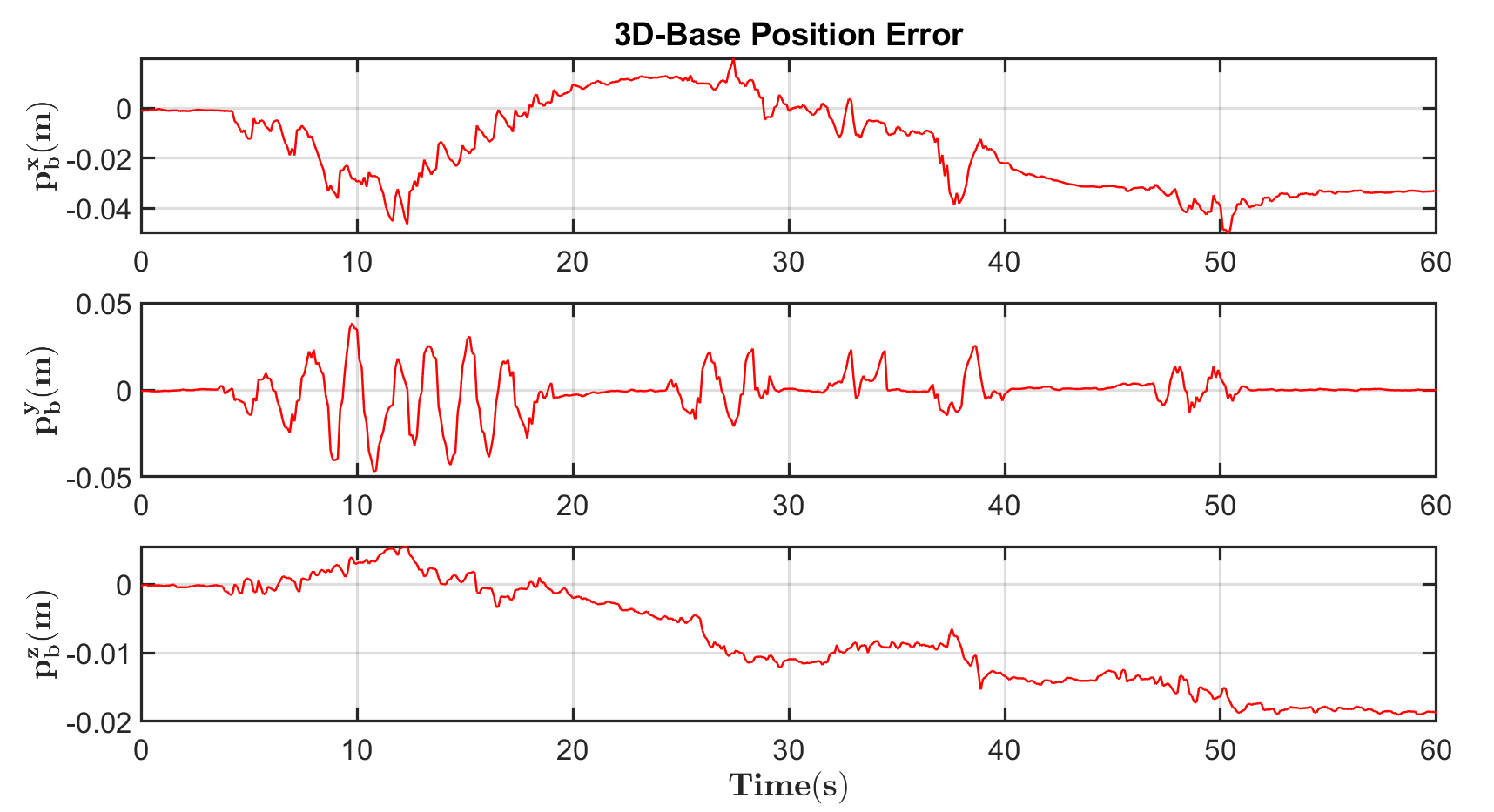}
    \caption{3D-Base position error of the estimated base position with SEROW from the corresponding ground-truth base position.}
    \label{fig:TALOS_SEROW}
\end{figure}

Finally, we employed LCD to predict the stable contact state probability for an actual TALOS humanoid and facilitate base state estimation with the State Estimation Robot Walking (SEROW) framework~\cite{PiperakisRAL}. The latter fuses effectively the contact state, kinematics, and the base IMU measurements to provide estimates for the base position, velocity and orientation. A vicon motion capture system was used to provide the ground-truth base pose every $200Hz$. Figure~\ref{fig:TALOS_SEROW} illustrates the 3D-base position error over time, whereby a slight drift is observed in the $x$ and $z$ axes for this $60s$ gait. The measured root mean square error  was particularly small, namely $0.0245m$, $0.0101m$, $0.0123m$ for the  base position and $0.7058deg$, $1.2035deg$, and $1.8426deg$ for the orientation, validating the employed stable contact state probabilities.

\subsection{Discussion}
We have demonstrated that an LCD model trained on a single dataset with the ATLAS robot walking over specific friction surfaces in RaiSim, achieves highly accurate contact detection. Additionally, the model generalizes well to contact estimation a) over surfaces with variable friction not previously included in the training dataset, b) with different robotic platforms scaling from small size light-weight robots such as NAO to full size heavy robots such as TALOS, and c) with different simulation platforms namely RaiSim and Gazebo. Consequently, it is rather straightforward to claim that the LCD architecture ought to have captured some robust contact features which are invariant to friction and to robot characteristics such as weight and height. Subsequently, we presented that the same architecture provides accurate contact estimation only with the GRF and the IMU data as input. The latter implies that this method can be readily adopted for robots with point feet, such as modern quadruped robots. These results pave the way for a holistic contact detection mechanism that is robot and contact agnostic.

\section{Conclusions}
\label{sec:conclusions}
In this article we introduced LCD,  a deep learning framework that provides a unified solution to contact detection by accurately and robustly estimating the leg contact state based solely on proprioceptive sensing. Although the latter rely on simulated ground-truth contact data for the training process, LCD generalizes across robotic platforms and can be readily transferred from simulation to real world setups. To reinforce further research endeavours we released LCD as an open-source ROS/Python package~\cite{LCD}.

LCD has been experimentally validated in terms of accuracy in simulation and has been compared against state-of-the-art approaches for contact detection with a simulated ATLAS, TALOS, and NAO robot. Additionally, its efficacy has been demonstrated in base estimation with an actual TALOS humanoid.

In future work, we aim at learning the contact state dynamics and utilize them to improve our state estimation and gait control schemes. Furthermore, we will investigate possible applications of the estimated contact state in humanoid visual SLAM~\cite{Hourdakis2021}.
  % This command serves to balance the column lengths
                                  % on the last page of the document manually. It shortens
                                  % the textheight of the last page by a suitable amount.
                                  % This command does not take effect until the next page
                                  % so it should come on the page before the last. Make
                                  % sure that you do not shorten the textheight too much.

%%%%%%%%%%%%%%%%%%%%%%%%%%%%%%%%%%%%%%%%%%%%%%%%%%%%%%%%%%%%%%%%%%%%%%%%%%%%%%%%

% NOTES ON PAPER (MIXALIS)
%
%   1) If the stable/unstable contact is predicted can we get an estimation about the friction coefficient of the ground?
%   2) F/T sensors are useless for estimating the quality of contact because the friction coefficient may vary. So the same values for Fx,Fy,Fz may be stable in some case and unstable in other cases
%      Others may use F/T because they are predicting contact or no contact but not actually the quality of the contact
%   3) When there are no bananas in the dataset. slippage occurs the moment  the foot touches the ground. --> Easy to predict with kNN / Decision Trees/ Random forest 
%       ^------------ Doesn't work for different friction coef. --> slip and stable data merge together 
%   4) 

%%%%%%%%%%%%%%%%%%%%%%%%%%%%%%%%%%%%%%%%%%%%%%%%%%%%%%%%%%%%%%%%%%%%%%%%%%%%%%%%%

%%%%%%%%%%%%%%%%%%%%%%%%%%%%%%%%%%%%%%%%%%%%%%%%%%%%%%%%%%%%%%%%%%%%%%%%%%%%%%%%
%%%%%%%%%%%%%%%%%%%%%%%%%%%%%%%%%%%%%%%%%%%%%%%%%%%%%%%%%%%%%%%%%%%%%%%%%%%%%%%%

\section*{Acknowledgement}
The authors would like to thank Olivier Stasse, LAAS-CNRS, for providing the necessary data for the actual TALOS experiment.

\bibliographystyle{IEEEtran}
\bibliography{root}

\end{document}